\documentclass[10pt,twocolumn,letterpaper]{article}

\usepackage[pagenumbers]{cvpr} 
\usepackage{listings}
\usepackage{xcolor}
\usepackage{multirow}
\usepackage{tabularx}
\lstdefinelanguage{json}{
    basicstyle=\small\ttfamily,
    string=[s]{"}{"},
    stringstyle=\color{blue!70!black},
    showstringspaces=false,
    breaklines=true
}
\definecolor{cvprblue}{rgb}{0.21,0.49,0.74}
\usepackage{colortbl}
\usepackage{xcolor}
\usepackage{float}
\usepackage{xfp}

\usepackage{multirow}
\newlength\savewidth\newcommand\shline{\noalign{\global\savewidth\arrayrulewidth
  \global\arrayrulewidth 1pt}\hline\noalign{\global\arrayrulewidth\savewidth}}
\newcommand{\tablestyle}[2]{\setlength{\tabcolsep}{#1}\renewcommand{\arraystretch}{#2}\centering\footnotesize}
\renewcommand{\paragraph}[1]{\vspace{1.25mm}\noindent\textbf{#1}}

\usepackage[pagebackref,breaklinks,colorlinks,allcolors=cvprblue]{hyperref}

\def\paperID{4119} 
\def\confName{CVPR}
\def\confYear{2026}

\title{ViMix-14M: A Curated Multi-Source Video–Text Dataset with Long-Form, High-Quality Captions and Crawl-Free Access}

\author{Timing Yang$^1$ \quad Sucheng Ren$^1$ \quad Alan Yuille$^1$ \quad Feng Wang$^{1}$\textsuperscript{*} \\
\\
$^1$Johns Hopkins University
}

\begin{document}

\twocolumn[{%
\renewcommand\twocolumn[1][]{#1}%
\maketitle
\vspace{-1cm}
\begin{center}
    \captionsetup{type=figure, width=\textwidth}
    \includegraphics[width=\textwidth]{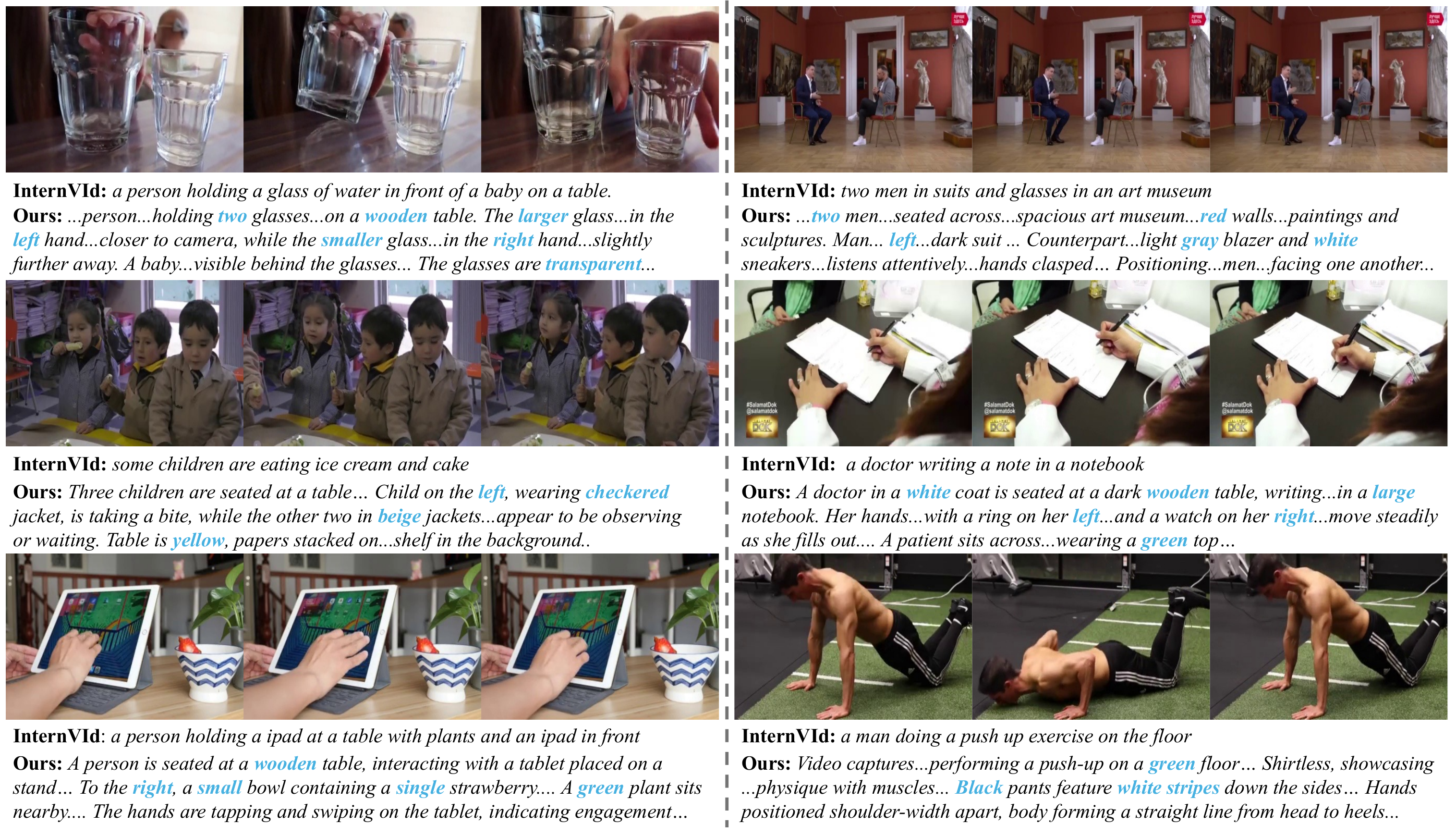}
    \caption{\textit{\textbf{Qualitative comparison of video captions.}} In contrast to existing datasets (\eg InternVid~\cite{wang2023internvid}) that use brief and generic descriptions, our ViMix-14M contains rich visual details in the captions. As shown, the text includes precise spatial relationship, object attributes, action states, and motion details, demonstrating superior scene understanding capabilities. Key visual details are marked in \textcolor{cyan}{blue}.}
    \label{fig:Comparison_six_samples}
\end{center}%

\vspace{+0.2cm}
}]

\maketitle
\begin{abstract}

\def\thefootnote{*}\footnotetext{Corresponding author. Email: \url{wangf3014@gmail.com}. Code is available at \url{https://github.com/yangtiming/ViMix-14M}.}

\vspace{-0.5cm}
Text-to-video generation has surged in interest since Sora~\cite{sora}, yet open-source models still face a data bottleneck: there is no large, high-quality, easily obtainable video–text corpus. Existing public datasets typically require manual YouTube crawling, which yields low usable volume due to link rot and access limits, and raises licensing uncertainty. This work addresses this challenge by introducing ViMix-14M, a curated multi-source video–text dataset of around 14 million pairs that provides crawl-free, download-ready access and long-form, high-quality captions tightly aligned to video. ViMix-14M is built by merging diverse open video sources, followed by unified de-duplication and quality filtering, and a multi-granularity, ground-truth-guided re-captioning pipeline that refines descriptions to better match actions, scenes, and temporal structure. We evaluate the dataset by multimodal retrieval, text-to-video generation, and video question answering tasks, observing consistent improvements over counterpart datasets. We hope this work can help removing the key barrier to training and fine-tuning open-source video foundation models, and provide insights of building high-quality and generalizable video-text datasets.

\end{abstract}    
\section{Introduction}
\label{sec:intro}

Recent advances in text-to-video generation~\cite{sora,wan2025wan,VideoLDM, yang2024cogvideox, bar2024lumiere} have demonstrated remarkable capabilities in producing photorealistic videos from natural language descriptions. These breakthroughs share a common foundation that the models are trained on massive-scale, high-quality video-text paired datasets with rich semantic annotations. The success mirrors earlier progress in text-to-image generation~\cite{balaji2022ediff,ramesh2021zero,Rombach_2022_CVPR,saharia2022photorealistic,yu2022scaling}, which transforms visual content creation through large-scale pre-training. However, while text-to-image generation has been democratized through large public datasets~\cite{schuhmann2022laion,sharma2018conceptual,changpinyo2021conceptual, kakaobrain2022coyo-700m, gadre2023datacomp,gadre2023datacomp}, video-language research faces a critical asymmetry. Closed-source commercial models leverage proprietary datasets with carefully curated video-text pairs, while open-source alternatives lack truly accessible, high-quality datasets. This fundamental accessibility barrier constrains the research community to develop competitive video understanding and generation systems, creating a persistent performance gap that has yet to be addressed through publicly available resources.

Current large-scale video datasets face three critical limitations that constrain model development. First, caption quality is fundamentally limited by ASR-based annotations. Most existing large-scale datasets rely on automatic speech recognition (ASR) to extract textual annotations from video audio~\cite{miech2019howto100m,xue2022hdvila,zellers2021merlot,lee2021acav100m}. For example, HowTo100M~\cite{miech2019howto100m} contains 136M video clips with ASR-derived transcripts, while HD-VILA-100M~\cite{xue2022hdvila} provides 103M clips with audio-based captions. These ASR transcripts describe the spoken narration rather than the visual content itself---capturing what is \emph{said} about the video rather than what is \emph{shown}. This audio-visual misalignment limits the visual semantic knowledge that models can acquire during training.

Second, limited scale and domain coverage restrict model generalization. Many existing datasets~\cite{soomro2012ucf101,kinetics-700,kay2017kinetics-400,kinetics-600,goyal2017something,ActivityNet,YouCook2,LSMDC}  concentrate on narrow domains with insufficient scale or diversity. For example, UCF101~\cite{soomro2012ucf101} contains only 13k videos across 101 action categories, while Kinetics-700~\cite{kinetics-700} focuses primarily on human actions with approximately 650k clips. Similarly, Something-Something V2~\cite{goyal2017something} emphasizes temporal reasoning through human-object interactions but remains constrained to 220k videos within a narrow domain of basic physical actions. Although these datasets provide clean categorical labels, their limited scale and narrow domain focus are insufficient for training large-scale video-language models that require diverse visual concepts and rich semantic supervision.

Third, dataset accessibility poses significant practical barriers. Many existing datasets source videos from YouTube~\cite{Webvid-10M,chen2024panda70m,xue2022hdvila,wang2023internvid}, where researchers face persistent download failures due to geographic access restrictions, content removal by uploaders, API rate limits, anti-scraping measures, and copyright takedowns. \textbf{\textit{As a result, the truly accessible videos of these datasets are often less than 1\% of the originally released or claimed volume.}} For example, WebVid-10M~\cite{Webvid-10M} has been permanently withdrawn following copyright holder requests. Panda-70M~\cite{chen2024panda70m} has been discontinued due to licensing restrictions, leaving only caption metadata accessible without corresponding videos. InternVid~\cite{wang2023internvid}, in contrast, offers a partially downloadable subset, yet the majority of videos still require YouTube access and remain subject to link rot, while its captions provide only brief descriptions that may miss temporal dynamics. This situation undermines reproducibility and democratization of video-language research, widening the performance gap between closed-source models with proprietary data and open-source alternatives.

We present ViMix-14M, a curated multi-source video-text dataset that addresses these fundamental limitations through three design principles.  First, we ensure reliable accessibility by providing datasets with stable hosting, clear licensing, and versioned releases, enabling reproducible use by the open-source community. Second, we prioritize caption quality through comprehensive re-captioning using Qwen2.5-VL-Instruct-7B~\cite{bai2025qwen2}. Leveraging the original ground-truth labels as prompts when available, we guide the model to generate rich, visually-grounded descriptions that capture objects, actions and scene dynamics. Beyond descriptive captions, we employ VBench~\cite{huang2023vbench} to systematically assess video quality through multi-dimensional evaluation, ensuring high standards and facilitating quality-based filtering for downstream applications. Third, we maximize content diversity by aggregating seven complementary sources spanning action recognition benchmarks, large-scale video collections, professional stock footage, and user-focused content.

Our dataset comprises 13.7 million videos totaling 22.8K hours, drawn from seven major sources: InternVid-10M-FLT\cite{wang2023internvid}, VideoUFO\cite{wang2025videoufo}, VidGen-1M\cite{tan2024vidgen}, Kinetics-700\cite{kinetics-700}, Something-Something V2\cite{goyal2017something}, OpenVideo\cite{openvideo}, and UCF-101\cite{soomro2012ucf101}.  To accommodate diverse application needs, we generate captions at three granularity levels: short captions for quick summarization, middle captions describing objects, colors, backgrounds, styles, and actions, and long captions that capture detailed spatial relationships, temporal reasoning, and comparative analysis between objects. Unlike existing datasets that provide only brief, generic descriptions lacking fine-grained visual information, our captions encode rich semantic details. As Figure~\ref{fig:Comparison_six_samples} demonstrates, our captions capture precise spatial relationships (\eg, ``larger glass in the left hand closer to camera"), specific color and material attributes (\eg, ``wooden table," ``checkered jacket"), exact quantities, and contextual scene elements. To ensure dataset video quality, we employ systematic VBench~\cite{huang2023vbench} evaluation across five dimensions: temporal flickering, subject consistency, background consistency, imaging quality, and aesthetic quality, guaranteeing high video standards across all samples. In summary, our work makes three principal contributions to advancing video-language research:

\begin{itemize}[leftmargin=*,itemsep=2pt,topsep=4pt]
\item We curate the largest stable video-language dataset comprising 13.7M videos with guaranteed crawl-free accessibility and clear licensing, eliminating the reproducibility barriers and legal ambiguities that have hindered open-source video-language model development.

\item We provide multi-granularity captions  that capture fine-grained visual details including precise spatial relationships, object attributes, color and material specifications, exact quantities, and semantic reasoning—significantly surpassing existing brief, generic descriptions.

\item We ensure content diversity through multi-source aggregation from seven major datasets and specialized domain collections, with systematic quality evaluation on temporal flickering, subject consistency, background consistency, imaging quality, and aesthetic quality to guarantee high video standards.
\end{itemize}

\section{Related Work}
\label{sec:RelatedWork}

\paragraph{Video-Language Datasets.} The success of image-language models~\cite{radford2021learning,jia2021scaling,li2022blip, liu2023visual} has been largely driven by large-scale image-text datasets~\cite{schuhmann2022laion, sharma2018conceptual, changpinyo2021conceptual, kakaobrain2022coyo-700m, gadre2023datacomp}
. However, scaling to video-language datasets presents unique challenges. Early video datasets such as UCF101~\cite{soomro2012ucf101} and Kinetics~\cite{kinetics-700} provide high-quality annotations but are limited in scale and primarily focus on action recognition with categorical labels rather than rich textual descriptions. To address scale limitations, several works~\cite{miech2019howto100m,xue2022hdvila,zellers2021merlot,Webvid-10M} leverage automatic speech recognition (ASR) to extract subtitles from videos, resulting in datasets like HowTo100M~\cite{miech2019howto100m} and HD-VILA-100M~\cite{xue2022hdvila}. While achieving scale, these ASR-based captions describe audio narration rather than visual content, creating an audio-visual misalignment that limits visual semantic learning. More recently, InternVid~\cite{wang2023internvid} provides 10M downloadable clips with generated captions averaging 17.6 words, though these descriptions may inadequately capture temporal dynamics due to frame-level captioning followed by summarization. Panda-70M~\cite{chen2024panda70m} employs multiple cross-modality teacher models to improve caption quality but faces video sources accessibility issues. In contrast, our dataset provides 13.7M stable, accessible videos with multi-granularity captions that capture rich visual semantics including spatial relationships, temporal dynamics, and object attributes.

\paragraph{Vision-Language Models.} Vision-language models learn correlations between visual inputs and textual representations, supporting downstream tasks such as text-guided image/video generation~\cite{balaji2022ediff,VideoLDM,ho2022imagen,ramesh2021zero,Rombach_2022_CVPR,saharia2022photorealistic,singer2022make,yu2022scaling}, captioning~\cite{alayrac2022flamingo,li2022blip,li2023blip2,liu2023visual}, VQA~\cite{chen2023minigpt,li2023videochat,zhu2023minigpt}, and retrieval~\cite{clip4clip, X-clip, li2023Unmaskedteacher,wang2023learning,wang2024sclip}.
Recent advances in large vision-language models (LVLMs) such as GPT-4V~\cite{achiam2023gpt4} and Gemini~\cite{team2023gemini} have demonstrated impressive multimodal understanding capabilities. For video-specific models, Video-LLaMA~\cite{Video-llama} and VideoChat~\cite{li2023videochat} extend LLMs to video understanding through learned visual projections. More recently, the Qwen series of models~\cite{bai2025qwen2,team2024qwen2,yang2025qwen3,wang2024qwen2}, including Qwen2.5-VL~\cite{bai2025qwen2}, have demonstrated strong performance on video understanding benchmarks. We leverage Qwen2.5-VL-Instruct-7B as our captioning model, which generates visually-grounded descriptions capturing objects, actions, and scene dynamics more effectively than frame-level or ASR-based approaches.

\paragraph{Video Annotation and Quality Evaluation.} Several concurrent works~\cite{wang2023internvid,chen2024panda70m,wang2023videofactory} explore automatic video annotation using vision-language models. InternVid~\cite{wang2023internvid} uses multi-modal inputs with LLM summarization, though LLMs may propagate errors from noisy captioning models. Panda-70M~\cite{chen2024panda70m} employs multiple teacher models including BLIP-2\cite{li2023blip2}, MiniGPT-4 \cite{Zhu2023MiniGPT4EV}, and video-specific models, demonstrating the benefits of multi-modal ensembling. Beyond annotation, ensuring video quality is critical for training robust models. We employ VBench~\cite{huang2023vbench} for systematic quality evaluation across temporal flickering, subject consistency, background consistency, imaging quality, and aesthetic quality. This comprehensive evaluation framework, combined with multi-granularity captioning at short, middle, and long levels, ensures our dataset provides both high-quality videos and rich semantic annotations suitable for training large-scale video-language models.

\section{Dataset ViMix-14M}
\label{sec:Dataset}

\begin{figure*}[t]
    \centering
    \vspace{-0.2cm}
    \includegraphics[width=1.0\linewidth]{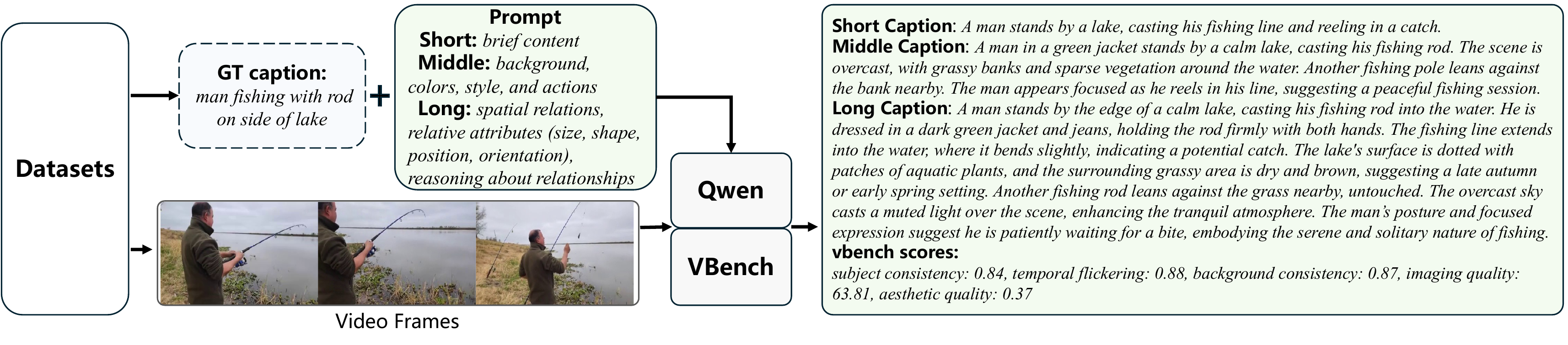}
    \caption{\textbf{\textit{Multi-granularity caption generation and filtering pipeline.}} Given video frames and ground truth captions (when available), we employ Qwen2.5-VL~\cite{bai2025qwen2} with three prompt levels to generate captions at varying granularities: short captions for brief content summarization, middle captions capturing scene details, and long captions encoding spatial relations and reasoning. Ground truth labels serve as optional contextual guidance to enhance caption quality but are not required for caption generation. VBench~\cite{huang2023vbench} evaluates video quality across five dimensions: subject consistency, temporal flickering, background consistency, imaging quality, and aesthetic quality. The fishing scene example demonstrates how caption granularity progressively captures richer visual and contextual details.}
    \label{fig:pipline}
    \vspace{-0.3cm}
\end{figure*}

Constructing Vimix includes three primary stages. First, we collect and curate videos from seven source domains under clear principles that all data must be crawl-free, directly downloadable and diverse across domains (detailed in Section~\ref{sec:data_collection}). This process yields a total of 13.7M curated videos, which are then subjected to captioning (Section~\ref{sec:captioning}) and filtering (Section~\ref{sec:quality}) stages. To best support diverse video–language tasks and broad applications, our data processing follows a customization-centric strategy. Specifically, for each video, we produce three caption granularities including a brief summary, a finer description, and a detailed, ground-truth-guided caption for reasoning-related tasks, which broadly cover diverse potential multimodal applications. For filtering, rather than discarding videos by fixed criteria, we provide rich per-video metadata, including duration, spatial size, and multiple video-quality indicators. This label-instead-of-filter strategy allows practitioners flexibly customize their own filtering rules by thresholding along specific dimensions when using ViMix-14M.

\subsection{Data Collection and Curation}
\label{sec:data_collection}

To ensure comprehensive coverage of visual concepts, temporal dynamics, and annotation granularities, we curate a diverse video corpus spanning seven complementary datasets. Our selection strategically combines four categories: (1) \textit{Action recognition benchmarks} (UCF-101~\cite{soomro2012ucf101}, Kinetics-700~\cite{kinetics-700}, Something-Something V2~\cite{goyal2017something}) providing foundational coverage of human activities and temporal reasoning, (2) \textit{Large-scale video-text datasets} (VidGen-1M~\cite{tan2024vidgen}, InternVid-10M-FLT~\cite{wang2023internvid}) featuring diverse real-world scenarios from user-generated content with detailed captions and high visual quality, (3) \textit{Professional stock footage} (OpenVideo~\cite{openvideo}) sourced from Pexels~\cite{pexels} curated video library, offering 106k+ professionally-produced clips with high production values that provide a complementary perspective to the spontaneous, naturalistic style of user-generated datasets, and (4) \textit{User-focused content} (VideoUFO~\cite{wang2025videoufo}) addressing real user interests in text-to-video generation with minimal overlap to existing datasets. This multi-source approach ensures balanced representation across action categories, scene types, motion patterns, and quality levels while covering both established benchmarks, professional cinematography, and contemporary user preferences.
All selected datasets are publicly available and can be efficiently accessed through major data platforms such as OpenDataLab~\cite{opendatalab} and HuggingFace~\cite{huggingface}, providing high-speed download infrastructure and streamlined data management. This accessibility facilitates reproducibility and adoption by the research community, eliminating the typical barriers associated with multi-source data collection.

\subsection{Multi-Granularity Captioning}
\label{sec:captioning}

We present our captioning and filtering pipeline in Figure~\ref{fig:pipline}. To obtain high-quality, visually-grounded captions that capture rich semantic information beyond ASR-based descriptions, we employ Qwen2.5-VL~\cite{bai2025qwen2}, a state-of-the-art open-source vision-language model with strong video understanding capabilities. The model processes video sequences at 1 FPS, enabling it to capture temporal dynamics and inter-frame relationships. Note that here we directly trust the captions produced by Qwen2.5-VL and our processing pipeline, treating the generated captions as ground truth, since we observe that their quality is already very close to human judgment. This assumption is supported by: 1) a randomly sampled batch evaluated by human annotators, where the pass rate exceeded 95\%; and 2) the detailed results in Section~\ref{sec:observation}, including direct caption quality analysis, text–video retrieval, video generation, and question answering, all of which indicate that ViMix-14M’s caption quality significantly outperforms existing counterparts.

To leverage existing annotations while generating richer descriptions, we integrate original ground-truth labels as contextual guidance when available. We prepend the original label to our prompts in the format: ``This video shows `[GT\_LABEL]'. [PROMPT]''. This approach provides the model with domain-specific context while allowing it to generate more detailed descriptions that extend beyond the concise ground-truth labels. For videos without ground-truth labels, we use the base prompts directly. To support diverse downstream applications with varying requirements for caption detail, we generate captions at three distinct granularity levels using the following prompts:

\begin{itemize}[leftmargin=*,itemsep=2pt,topsep=2pt]
\item Short: ``\textit{Briefly describe the main content of the video in no more than 20 words.}''

\item Middle: ``\textit{Describe the objects in the video, including their colors, background, style, and actions, in 40 to 60 words.}''

\item Long: ``\textit{Provide a detailed caption that not only describes the objects and actions in the video but also highlights their relative attributes, such as differences in size, shape, height, body build, orientation, position, or quantity. The caption should also include reasoning that explains relationships between the objects, movements, or events, going beyond surface-level description, in 80 to 130 words}''
\end{itemize}

\begin{table*}[!t]
\centering
\vspace{-0.2cm}
\tablestyle{5pt}{1.1}
\begin{tabular}{l|ccc|ccc|ccccc}
\multirow{2}{*}{Dataset} & \multicolumn{3}{c|}{Statistics} & \multicolumn{3}{c|}{Duration (words)} & \multicolumn{5}{c}{Quality Metrics} \\
& \#Videos & Total hrs. & Avg. len.(s) & Short & Middle & Long & Subject & Background & Temporal & Aesthetic & Imaging \\
& & & & & & & consist. & consist. & flickering & quality & quality \\
\shline
InternVId & 10.6M & 16.3K & 5.5 & 14.4 & 50.5 & 109.2 & 0.87 & 0.91 & 0.93 & 0.41 & 57.50 \\
VideoUFO & 1.09M & 2.12K & 7.0 & 15.9 & 53.2 & 113.4 & 0.80 & 0.86 & 0.91 & 0.26 & 52.25 \\
VidGen-1M & 1.00M & 2.26K & 8.1 & 13.7 & 47.2 & 110.1 & 0.83 & 0.90 & 0.87 & 0.47 & 63.39 \\
Kinetics-700 & 0.63M & 1.58K & 9.0 & 13.6 & 52.2 & 111.4 & 0.79 & 0.89 & 0.87 & 0.43 & 52.47 \\
Sthv2 & 0.22M & 234 & 3.8 & 13.9 & 46.1 & 107.8 & 0.85 & 0.90 & 0.88 & 0.33 & 54.04 \\
OpenVideo & 0.11M & 331 & 11.2 & 17.8 & 57.4 & 117.8 & 0.90 & 0.92 & 0.91 & 0.53 & 68.87 \\
UCF-101 & 0.01M & 26.6 & 7.2 & 13.4 & 50.3 & 109.3 & 0.88 & 0.92 & 0.92 & 0.40 & 45.79 \\

\rowcolor{cvprblue!15}
\textbf{ViMix-14M} & \textbf{13.7M} & \textbf{22.8K} & \textbf{6.0} & \textbf{14.4} & \textbf{50.6} & \textbf{109.8} & \textbf{0.86} & \textbf{0.90} & \textbf{0.92} & \textbf{0.40} & \textbf{57.33} \\
\end{tabular}
\caption{\textbf{\textit{ViMix-14M Dataset composition}} with video counts, durations, caption statistics, and quality metrics.}
\label{tab:dataset_stats}
\vspace{-0.3cm}
\end{table*}

\begin{figure}[h]
    \centering
    \includegraphics[width=1\linewidth]{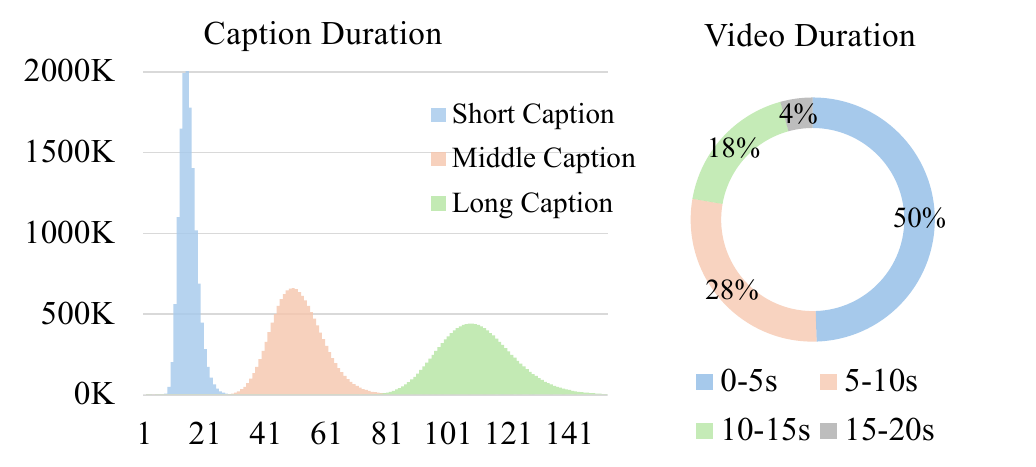}
    \caption{\textbf{\textit{Caption and video statistics.}} Left: frequency of captions (y-axis) by word count (x-axis) for short, middle, and long granularities. Right: percentage distribution of video durations.}
    \label{fig:duration_dist}
  
\end{figure}

\subsection{Video Quality Assessment}
\label{sec:quality}

To ensure high video quality standards across our dataset, we employ VBench~\cite{huang2023vbench}, a comprehensive video quality evaluation framework that assesses videos across multiple dimensions with alignment to human perception. This quality assessment allows us to filter low-quality videos and provide quality scores that enable researchers to customize their training data according to specific requirements. In detail, VBench provides a multi-dimensional evaluation that decomposes video generation quality into specific, hierarchical dimensions including temporal quality, frame-wise quality, and video-condition consistency. It goes beyond single-metric scoring by offering automated, multi-dimensional evaluation, with its metrics validated against human preference data. We evaluate each video across five key dimensions in VBench:

\begin{itemize}[leftmargin=*,itemsep=2pt,topsep=2pt]
\item \textit{Temporal Flickering} assesses local temporal inconsistencies in brightness and color across consecutive frames. Lower flickering scores indicate better stability, which is essential for training models that learn temporal patterns.
\item \textit{Subject Consistency} evaluates whether the main subject's appearance and identity remain stable throughout the video duration. This metric is critical for learning robust object representations across temporal sequences.
\item \textit{Background Consistency} measures the temporal stability of background elements and scene context. Stable backgrounds enable models to better distinguish foreground dynamics from static environmental features.
\item \textit{Aesthetic Quality} assesses the visual appeal and artistic value of video frames, including compositional balance, color harmony, and photorealistic rendering quality.
\item \textit{Imaging Quality} evaluates technical clarity aspects such as sharpness, proper exposure, color fidelity, and the absence of visual artifacts like blur or noise.
\end{itemize}

\noindent\textbf{Data Structure.}
We organize our curated corpus into standardized JSON for each source dataset. Each video entry follows a consistent schema:

\begin{lstlisting}[
  language=json,
  basicstyle=\footnotesize\ttfamily,
  frame=single,
  backgroundcolor=\color{gray!3},
  rulecolor=\color{gray!25},
  stringstyle=\color{cvprblue!80!},
  showstringspaces=false,
  xleftmargin=8pt,
  xrightmargin=8pt,
  aboveskip=8pt,
  belowskip=8pt
]
{
  "segment_id": <unique video identifier>,
  "dataset_name": <source dataset name>,
  "split": <train|val|test>,
  "duration": <video length in seconds>,
  "use_gt_label": <True|False>,
  "start_time": <timestamp in HH:MM:SS.mmm>,
  "end_time": <timestamp in HH:MM:SS.mmm>,
  "caption_short_en": <10-20 words>,
  "caption_middle_en": <40-60 words>,
  "caption_long_en": <80-130 words>,
  "vbench_scores": {
    "subject_consistency": <0-1 score>,
    "background_consistency": <0-1 score>,
    "temporal_flickering": <0-1 score>,
    "aesthetic_quality": <0-1 score>,
    "imaging_quality": <absolute score>
  }
}
\end{lstlisting}

The \texttt{use\_gt\_label} field indicates whether original ground-truth labels are retained from source datasets. Temporal boundaries (\texttt{start\_time}, \texttt{end\_time}) represent actual video segment positions, where \texttt{end\_time - start\_time = duration} always holds. This structure enables efficient data loading, granularity-specific caption retrieval, and quality-based filtering.

\noindent\textbf{Dataset Statistics.} Table~\ref{tab:dataset_stats} presents statistics across all seven source datasets. Our corpus comprises 13.7M video clips totaling 22.8K hours, with an average duration of 6.0 seconds. Figure~\ref{fig:duration_dist} illustrates the distribution of caption lengths and video durations across our dataset. The left histogram displays three distinct peaks corresponding to our multi-granularity captions: short captions averaging 14.4 words, middle captions at 50.6 words, and long captions at 109.8 words. The right pie chart shows that nearly half (50\%) of videos are 0-5 seconds, with 28\% in the 5-10 second range, 18\% at 10-15 seconds, and 4\% at 15-20 seconds.

\section{Observations}
\label{sec:observation}

\subsection{Multi-Dimensional Caption Quality Analysis}

\begin{figure}[h!]
    \centering
    \includegraphics[width=1\linewidth]{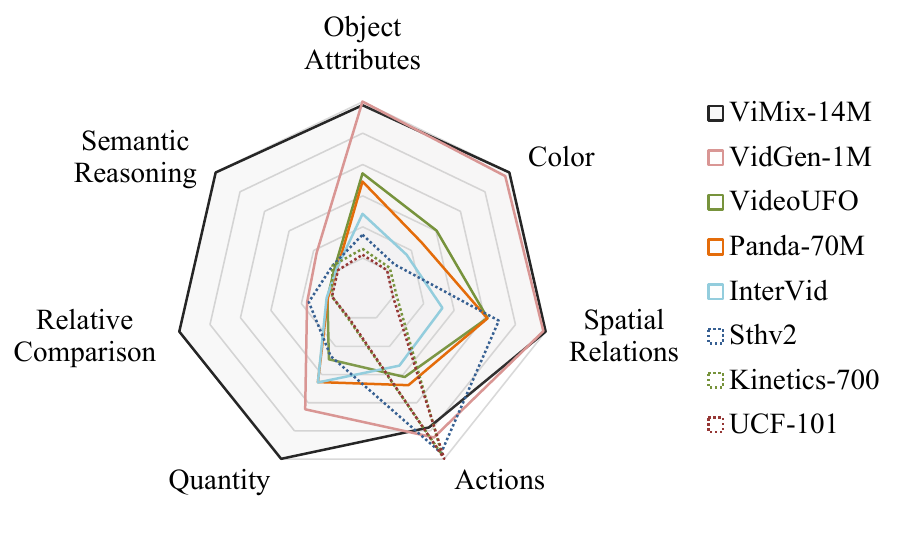}
    \caption{\textbf{\textit{Multi-dimensional caption quality comparison.}} Our captions demonstrate superior semantic richness across all evaluated dimensions compared to existing datasets.}
    \label{fig:radar}
\vspace{-0.4cm}
\end{figure}

To assess caption semantic richness, we conduct a multi-dimensional evaluation across seven key dimensions of visual understanding. Specifically, \textit{Object Attributes} examines physical and visual characteristics including shape, size, material, texture, and structure. \textit{Spatial Relations} evaluates spatial organization through position, direction, distance, and relational patterns. \textit{Color} assesses color-related information including specific hues, tones, combinations, and contrasts. \textit{Quantity} measures numerical information through exact counts, quantity words, distribution, and groups. \textit{Relative Comparison} evaluates explicit comparisons between objects through size differences, contrast words, and relative terms. \textit{Actions} examines movements and behaviors through action verbs, manner descriptions, and object interactions. \textit{Semantic Reasoning} assesses high-level interpretation beyond surface description including symbolism, inference, atmosphere, and abstract concepts.
Using Qwen3-4B-Instruct \cite{yang2025qwen3} as an automated evaluator, we analyze captions through dimension-specific binary classification prompts. Each prompt begins with a yes/no question in the form \textit{``Does this caption ...?''}, followed by key cues, example pairs, and an instruction to answer with ``True'' or ``False.''  Figure~\ref{fig:radar} presents the comparative analysis across datasets. Our captions demonstrate substantially superior coverage across all dimensions. While action recognition datasets like Kinetics-700 and UCF-101 excel in Actions, our multi-granularity approach achieves balanced and comprehensive coverage spanning low-level perceptual details to high-level semantic reasoning.

\begin{figure*}[t!]
\centering
\vspace{-0.2cm}
\includegraphics[width=\textwidth]{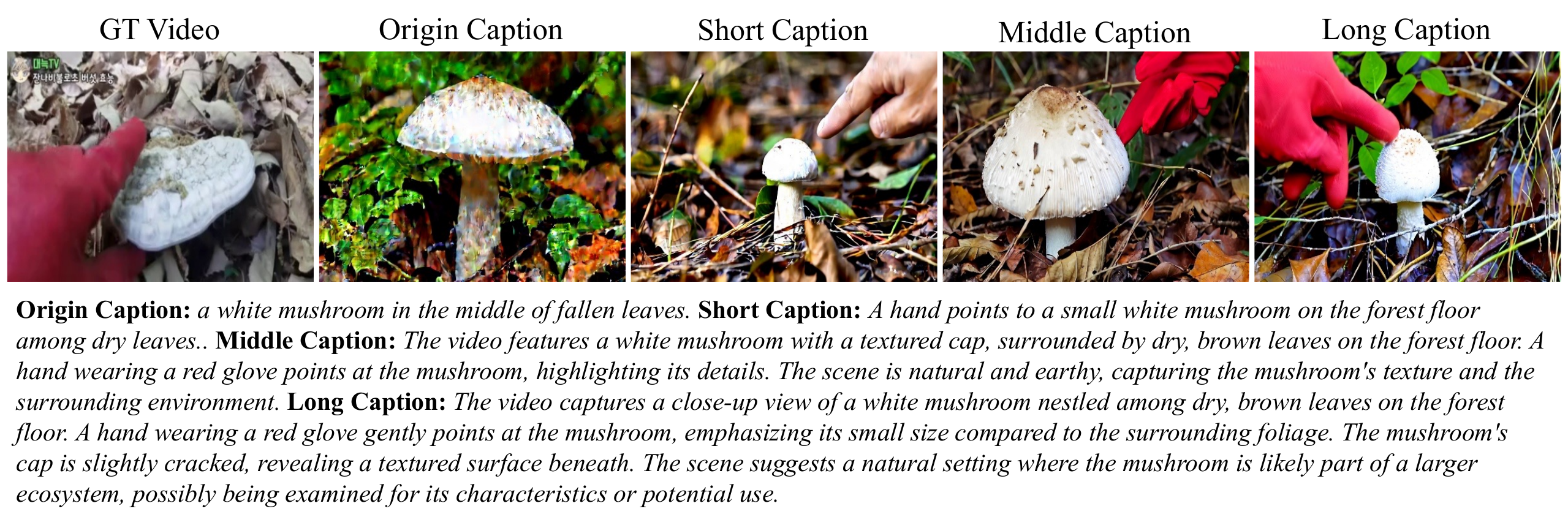}
\vspace{-0.75cm}
\caption{\textbf{\textit{Video generation quality comparison across caption granularities.}} Origin Caption (\eg InternVid~\cite{wang2023internvid}) provides the basic scene; Short Caption adds hand-mushroom interaction; Middle Caption correctly predicts the gloved hand with color; Long Caption achieves the best results with improved hand posture and correct contact direction. Quality hierarchy: Origin $<$ Short $<$ Middle $<$ Long.}
\label{fig:Generation_vis}
\end{figure*}

\begin{table*}[t]
\centering
\tablestyle{4pt}{1.0}
\small
\begin{tabular}{l|ccccccc|ccccccc}
\multirow{2}{*}{Caption} & \multicolumn{7}{c|}{Text-to-Video} & \multicolumn{7}{c}{Video-to-Text} \\
& InternVid & K700 & UCF & VidGen & UFO & Open & Sthv2 & InternVid & K700 & UCF & VidGen & UFO & Open & Sthv2 \\
\shline
\multicolumn{15}{l}{\it \textcolor{gray}{Recall@1 (\%) ↑}} \\
Origin & 76.5 & 31.0 & 12.4 & 93.0 & 65.4 & 89.8 & 22.4 & 73.5 & 34.0 & 12.6 & 93.6 & 71.2 & 90.8 & 24.0 \\

\multicolumn{15}{l}{\it \textcolor{cvprblue!60}{ViMix-14M (results shown below):}} \\
\rowcolor{cvprblue!5}
Short & 83.6 & 71.8 & 39.8 & 87.2 & 82.0 & 88.2 & 47.8 & 80.0 & 75.8 & 41.4 & 87.4 & 81.0 & 90.4 & 50.6 \\
\rowcolor{cvprblue!10}
Middle & \textbf{90.2} & 84.2 & \textbf{63.4} & \textbf{95.6} & \textbf{91.4} & 88.8 & 61.8 & \textbf{89.8} & \textbf{90.8} & \textbf{68.0} & 94.4 & 91.8 & 90.4 & 69.0 \\
\rowcolor{cvprblue!20}
Long & 89.6 & \textbf{86.6} & 59.6 & 94.2 & 88.6 & \textbf{89.8} & \textbf{66.0} & 87.2 & 89.6 & 64.4 & \textbf{95.8} & \textbf{92.6} & \textbf{91.2} & \textbf{72.4} \\
\hline
\multicolumn{15}{l}{\it \textcolor{gray}{Recall@5 (\%) ↑}} \\
Origin & 95.0 & 58.4 & 52.2 & 99.4 & 87.4 & 99.4 & 45.4 & 91.0 & 62.0 & 53.0 & 99.6 & 91.4 & 99.6 & 47.5 \\

\multicolumn{15}{l}{\it \textcolor{cvprblue!60}{ViMix-14M (results shown below):}} \\
\rowcolor{cvprblue!5}
Short & 95.8 & 92.8 & 81.0 & 99.0 & 94.6 & 98.2 & 72.0 & 93.4 & 95.8 & 85.0 & 99.2 & 94.4 & 99.2 & 76.6 \\
\rowcolor{cvprblue!10}
Middle & \textbf{98.4} & 96.4 & \textbf{92.8} & 99.8 & \textbf{98.4} & 99.0 & 84.2 & \textbf{96.8} & 97.2 & 94.2 & \textbf{100.0} & 99.2 & 99.2 & 88.8 \\
\rowcolor{cvprblue!20}
Long & 98.2 & \textbf{97.0} & 91.4 & \textbf{100.0} & 97.6 & \textbf{99.4} & \textbf{86.2} & 96.0 & \textbf{98.4} & 93.0 & \textbf{100.0} & \textbf{98.6} & \textbf{99.6} & \textbf{91.0} \\
\hline
\multicolumn{15}{l}{\it \textcolor{gray}{Mean Rank ↓}} \\
Origin & 2.3 & 39.5 & 16.2 & 1.1 & 4.6 & 1.2 & 43.5 & 3.1 & 28.6 & 19.3 & 1.1 & 3.5 & 1.2 & 36.3 \\

\multicolumn{15}{l}{\it \textcolor{cvprblue!60}{ViMix-14M (results shown below):}} \\
\rowcolor{cvprblue!5}
Short & 2.0 & 2.5 & 4.3 & 1.3 & 2.2 & 1.3 & 12.5 & 2.5 & 2.3 & 3.4 & 1.2 & 2.5 & 1.2 & 9.3 \\
\rowcolor{cvprblue!10}
Middle & \textbf{1.3} & 1.8 & \textbf{2.2} & \textbf{1.1} & \textbf{1.4} & 1.3 & 5.0 & \textbf{1.5} & 1.5 & \textbf{1.9} & 1.1 & 1.4 & 1.2 & 3.8 \\
\rowcolor{cvprblue!20}
Long & 1.3 & \textbf{1.8} & 2.6 & 1.1 & 1.5 & \textbf{1.2} & \textbf{4.5} & 1.7 & \textbf{1.5} & 2.0 & \textbf{1.0} & \textbf{1.3} & \textbf{1.2} & \textbf{3.6} \\
\end{tabular}
\caption{\textbf{\textit{Bidirectional video-text retrieval across different caption lengths.}} We compare four caption configurations on seven datasets using Recall@1, Recall@5, and Mean Rank. \textbf{Bold} indicates best performance per dataset. Shading from \colorbox{cvprblue!5}{light} to \colorbox{cvprblue!10}{medium} to \colorbox{cvprblue!20}{dark} corresponds to caption length from Short to Middle to Long.}
\label{tab:retrieval}
\vspace{-0.2cm}
\end{table*}

\subsection{Text-Video Retrieval Evaluation}

To validate caption quality, we conduct retrieval experiments using Clip4Clip~\cite{clip4clip} on 500 randomly sampled video-text pairs per dataset. We evaluate both text-to-video (T2V) and video-to-text (V2T) retrieval, comparing four text representations: original captions, short captions, middle captions, and long captions. 
Results in Table~\ref{tab:retrieval} demonstrate that our generated captions consistently and substantially outperform origin labels across both directions. For T2V retrieval, improvements are significant on action recognition datasets where origin labels provide limited semantic information: middle captions achieve 63.4\% R@1 on UCF-101 \emph{vs.} 12.4\% for origin labels, 86.6\% on Kinetics-700 \emph{vs.} 31.0\%, and 66.0\% on Something-Something V2 \emph{vs.} 22.4\%. On datasets with richer visual content, middle captions achieve T2V performance exceeding 90\% R@1: 95.6\% on VidGen-1M, 91.4\% on VideoUFO, and 90.2\% on InternVid-10M-FLT.
For V2T retrieval, long captions demonstrate superior or competitive performance: 95.8\% R@1 on VidGen-1M, 92.6\% on VideoUFO, and 72.4\% on Something-Something V2, suggesting that richer textual descriptions provide better discriminative power when selecting among multiple text candidates. The improvements across both directions validate that our visually-grounded, multi-granularity captions establish robust semantic alignment between visual and textual modalities, supporting diverse downstream retrieval applications.

\begin{figure*}[h]
    \centering
    \vspace{-0.3cm}
    \includegraphics[width=1\linewidth]{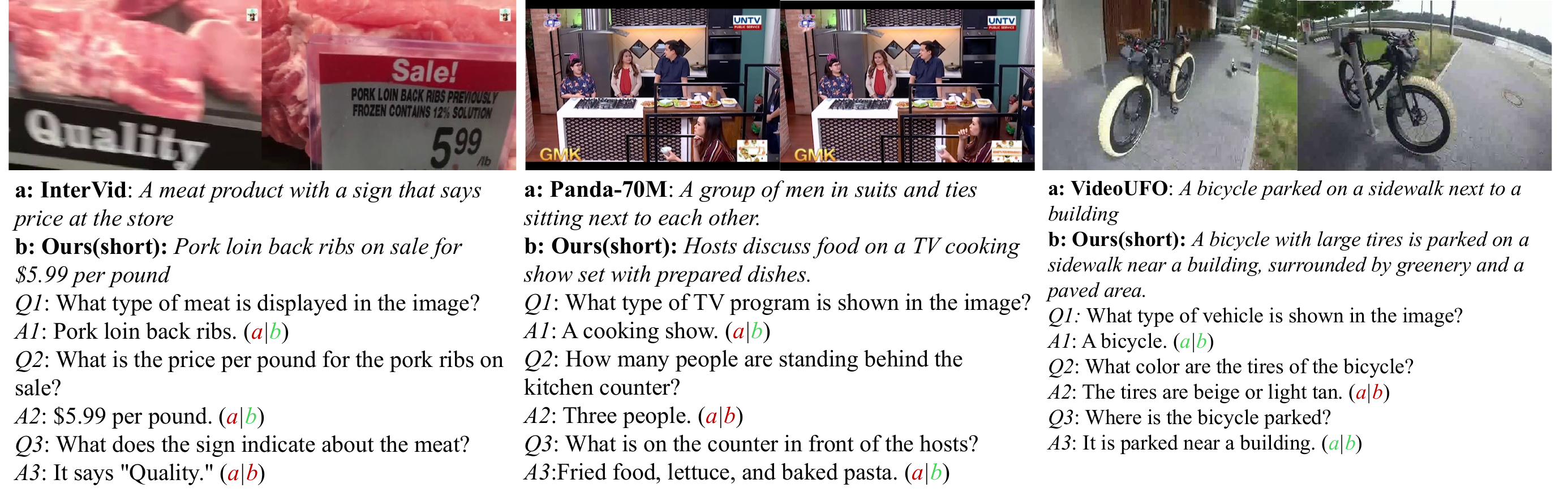}
    \vspace{-0.6cm}
    \caption{\textbf{\textit{Caption evaluation via video question-answering tasks.}} Comparison between baseline captions (a) and ours (b) across multiple datasets. Color markers show answer coverage. Our captions capture richer visual details, leading to more accurate responses for attribute-specific and content-focused questions.}
    \label{fig:VQA}
    \vspace{-0.3cm}
\end{figure*}

\subsection{Quality Evaluation via Video Generation}

To evaluate caption quality for text-to-video generation, we measure CLIP-based \cite{clip} similarity scores between videos generated by Wan2.2 \cite{wan2025wan} model from different captions and their corresponding source videos using Clip4Clip~\cite{clip4clip}. Table~\ref{tab:caption_similarity} demonstrates that our generated captions achieve substantially higher video-text alignment scores compared to original labels, with long captions yielding the highest average similarity (0.70) and middle captions achieving 0.69, both significantly exceeding the origin baseline (0.56). Action recognition datasets exhibit the most significant improvements: \eg, UCF-101 increases from 0.42 to 0.74 (middle, +0.32), and Kinetics-700 rises from 0.45 to 0.70 (middle, +0.25). Long captions achieve superior alignment in most datasets such as VideoUFO and InternVid. High-quality source datasets show strong alignment, with OpenVideo achieving the highest score (0.80 for long captions) and VidGen-1M following with 0.70. These results validate that our multi-granularity captions establish strong video-text alignment, supporting text-to-video generation and video-language tasks.

As shown in Figure~\ref{fig:Generation_vis}, captions evolve from basic object descriptions (``a white mushroom in fallen leaves") to progressively richer representations incorporating human actions (``hand points"), fine-grained attributes (``textured cap", ``red glove"), comparative spatial relationships (``small size compared to surrounding foliage"), and contextual reasoning (``examined for characteristics"). This progression reflects comprehensive visual understanding, capturing not only objects but also their properties, spatial arrangements, and context.

\subsection{Quality Assessment via VQA}

To further validate the semantic richness of our generated captions, we conduct text-only Video Question Answering (VQA) experiments where models answer video-related questions using solely textual descriptions without visual input. We evaluate caption quality by comparing the accuracy of answers derived from original captions \emph{vs.} our captions across diverse question types, including object counting, object identification, attribute recognition (e.g., color, price), spatial localization, and scene understanding.
For each video sample, we generate 3 diverse questions covering various semantic aspects using GPT-5 \cite{openai2025gpt5}, which views the original video frames and creates question-answer pairs as ground truth. We then employ GPT-5 to answer these questions based purely on textual captions—both original and ours—without access to visual content. Answer correctness is evaluated using an LLM-as-judge approach, where GPT-5 assesses semantic equivalence between predicted and ground truth answers, accounting for synonyms and paraphrasing.

Figure~\ref{fig:VQA} illustrates qualitative comparisons across multiple datasets. For baseline captions (a), the sparse descriptions limit answer coverage, particularly for attribute-specific questions (\eg, ``What color is the object?") and fine-grained content queries (\eg, ``How many people are visible?"). In contrast, our captions (b) provide richer visual details—capturing object counts, colors, spatial arrangements, and contextual information—enabling accurate responses across diverse question types. The color-coded markers demonstrate that our captions achieve comprehensive answer coverage, while baseline captions fail to provide sufficient information for precise answers. These results demonstrate that our visually-grounded multi-granularity captions encode richer semantic information compared to existing annotations, enabling effective video understanding through text alone and validating their utility for caption-based video reasoning tasks.

\begin{table}[t]
\centering
\tablestyle{3pt}{1.1}
\begin{tabular}{l|ccccccc|c}
Caption & Sthv2 & VidGen & UFO & Open & InternVid & UCF & K700 & Avg \\
\shline
Origin & 0.50 & 0.69 & 0.55 & 0.79 & 0.54 & 0.42 & 0.45 & 0.56 \\
\multicolumn{9}{l}{\it \textcolor{cvprblue!60}{ViMix-14M (results shown below):}} \\
\rowcolor{cvprblue!5}
Short & 0.58 & 0.60 & 0.57 & 0.69 & 0.55 & 0.64 & 0.64 & 0.61 \\
\rowcolor{cvprblue!10}
Middle & 0.65 & 0.70 & 0.59 & 0.77 & 0.66 & \textbf{0.74} & \textbf{0.71} & 0.69 \\
\rowcolor{cvprblue!20}
Long & \textbf{0.65} & \textbf{0.70} & \textbf{0.62} & \textbf{0.80} & \textbf{0.70} & 0.72 & 0.70 & \textbf{0.70} \\
\end{tabular}
\caption{\textbf{\textit{Video-text similarity}} scores across caption granularities (mean CLIP similarity between generated and source videos).}
\label{tab:caption_similarity}
\vspace{-0.3cm}
\end{table}

\subsection{Video Quality Analysis via VBench}

\begin{table}[t]
\centering
\tablestyle{4pt}{1.1}
\vspace{-0.4cm}
\begin{tabular}{c|ccc|ccc}
\multirow{2}{*}{Threshold} & \multicolumn{3}{c|}{T2V R@1 ↑} & \multicolumn{3}{c}{V2T R@1 ↑} \\
& Short & Middle & Long & Short & Middle & Long \\
\shline
Baseline & 83.6 & 90.2 & 89.6 & 80.0 & 89.8 & 87.2 \\
subject\_cons. & 83.0 & 91.0 & 89.6 & 82.0 & 91.0 & 89.6 \\
temporal\_flick. & 82.4 & 91.4 & 91.0 & 80.4 & 91.6 & 89.6 \\
background\_cons. & 82.2 & 89.6 & 89.8 & 79.2 & 91.2 & 90.4 \\
aesthetic\_qual. & 83.0 & 91.4 & 90.4 & \textbf{84.0} & 92.0 & 90.6 \\
\rowcolor{cvprblue!20}
imaging\_qual. & \textbf{88.4} & \textbf{94.2} & \textbf{93.8} & 83.2 & \textbf{93.6} & \textbf{92.2} \\
\end{tabular}
\caption{\textbf{\textit{Comparison of R@1 scores}} for T2V and V2T retrieval across different VBench quality dimensions. Imaging quality filtering (highlighted) achieves superior performance.}
\label{tab:r1_comparison}
\vspace{-0.3cm}
\end{table}

 we examine how different VBench quality thresholds affect retrieval performance. We apply quality-based filtering with thresholds of 50 for imaging quality, 0.3 for aesthetic quality, 0.7 for background consistency, 0.7 for temporal flickering, and 0.5 for subject consistency, then evaluate using Clip4Clip~\cite{clip4clip}. Table~\ref{tab:r1_comparison} shows that imaging quality filtering achieves the best overall performance, with T2V R@1 scores of 88.4\%, 94.2\%, and 93.8\% for short, medium, and long videos—improvements of 4.8, 4.0, and 4.2 points over baseline. For V2T retrieval, it attains 93.6\% and 92.2\% on medium and long videos. Notably, aesthetic quality performs best on short V2T tasks (84.0\%), suggesting that aesthetic features better capture semantic information in brief content. These results demonstrate that quality-based filtering effectively enhances cross-modal retrieval, with imaging quality being the most impactful dimension.

\section{Conclusion}
\label{sec:conclusion}

We present ViMix-14M, a large-scale video-text dataset comprising 13.7 million videos that addresses fundamental limitations in video-language research through crawl-free accessibility, multi-granularity high-quality captions, and systematic quality evaluation. Through comprehensive re-captioning using Qwen and multi-source aggregation from seven complementary datasets, our dataset achieves substantial improvements across multimodal retrieval, text-to-video generation, and video question answering tasks. Our work provides a stable, legally clear, and semantically rich resource that enables the broader research community to develop competitive video-language models.
{
    \small
    \bibliographystyle{ieeenat_fullname}
    \bibliography{main}
}

\clearpage
\setcounter{page}{1}
\maketitlesupplementary

\subsection*{A. Video Compression for Quality Assessment}
\begin{figure}[h!]
    \centering
    \includegraphics[width=0.95\linewidth]{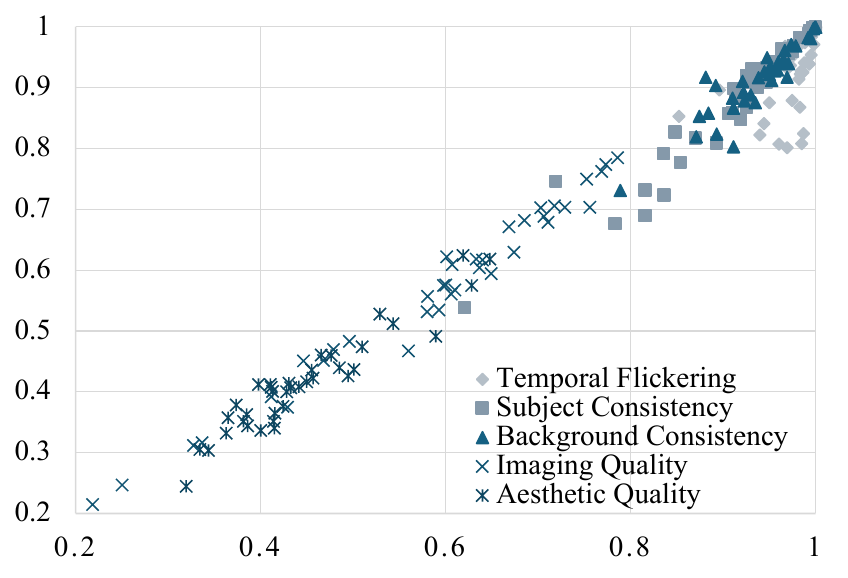}
     \captionsetup{width=0.95\linewidth}
    \caption{\textbf{Impact of video compression on VBench quality metrics.}
    X-axis: original scores; Y-axis: compressed scores. Imaging quality is normalized to 0-1. Strong correlation indicates minimal compression impact.}
    \label{fig:vbench_quality}
\end{figure}

To enable efficient processing of large-scale datasets, we compress videos to 10 frames per video before VBench quality assessment. Figure~\ref{fig:vbench_quality} shows strong correlation between original and compressed quality scores across all five dimensions, validating that this compression preserves evaluation reliability.

\subsection*{B. Implementation Details.}
 We process videos at 1 FPS and set maximum token limits of 30, 100, and 200 for short, middle, and long captions respectively. For videos longer than 20 seconds, we randomly sample a 5-15 second segment to balance computational cost and temporal information preservation. For video quality assessment, we sample up to 10 frames per video; videos with fewer than 5 frames are excluded from the dataset.  We deploy the captioning pipeline across 36 NVIDIA A5000 GPUs with batch processing, where each video is processed through all three prompt levels simultaneously using batch inference with mixed precision (FP16) to accelerate generation. The complete pipeline, including caption generation and video quality assessment, takes approximately one month to process the entire dataset. For CLIP4CLIP, we use the pre-trained checkpoint from HuggingFace (\textit{Searchium-ai/clip4clip-webvid150K}).

\subsection*{C. Additional Video Generation Examples}

\begin{figure*}[!t]
    \centering
    \includegraphics[width=0.95\linewidth]{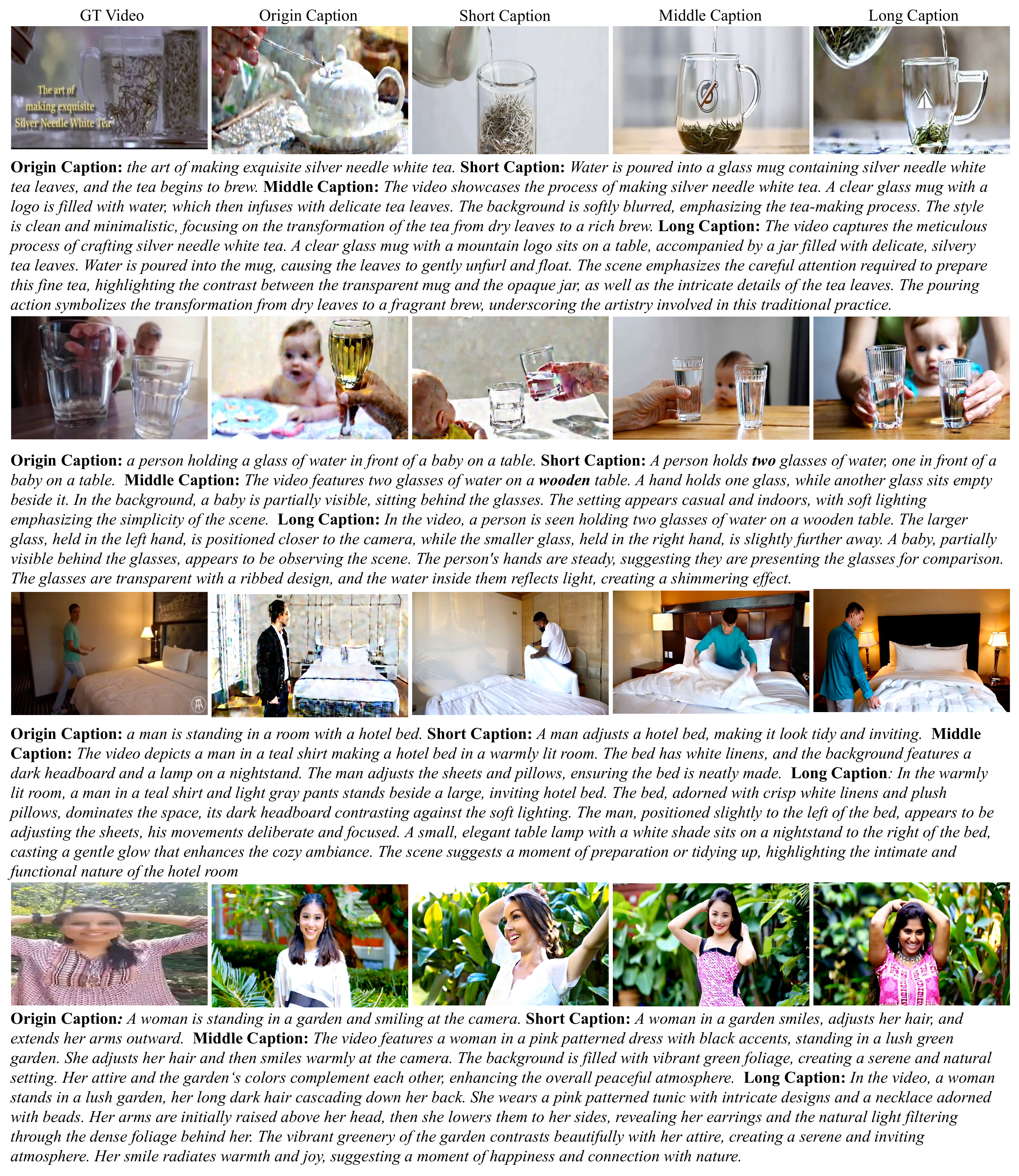}
    \captionsetup{width=0.95\linewidth}
    \caption{\textbf{Additional video generation examples across caption granularities.} Similar to Figure in the main paper, these examples demonstrate how increasing caption detail (from Origin to Short to Middle to Long) progressively improves generation quality. Each row shows: GT Video, Origin Caption generation, Short Caption generation, Middle Caption generation, and Long Caption generation, with the corresponding captions displayed below.}
    \label{fig:supp_gen}
\end{figure*}

Figure~\ref{fig:supp_gen} presents additional examples demonstrating how caption granularity progressively improves video generation quality, from origin captions providing basic context, to short captions adding primary actions, middle captions incorporating visual details like colors and spatial relationships, and long captions achieving the highest quality through fine-grained attributes and contextual reasoning.

Across diverse scenarios—from tea preparation and childcare to hotel room settings and outdoor garden activities—we observe consistent quality improvements as caption length and detail increase. For example, the tea preparation scene demonstrates how long captions describing material properties ("transparent mug" versus "opaque jar"), motion dynamics ("leaves gently unfurl and float"), and symbolic meaning enable generation of videos with correct attributes and appropriate atmosphere. Similarly, the hotel bed example shows how descriptions of lighting, positioning, and ambiance translate directly to generation quality, with long captions producing videos that accurately capture color schemes, spatial layouts, and mood.

\subsection*{D. Additional VQA Examples}

\begin{figure*}[!t]
    \centering
    \includegraphics[width=0.95\linewidth]{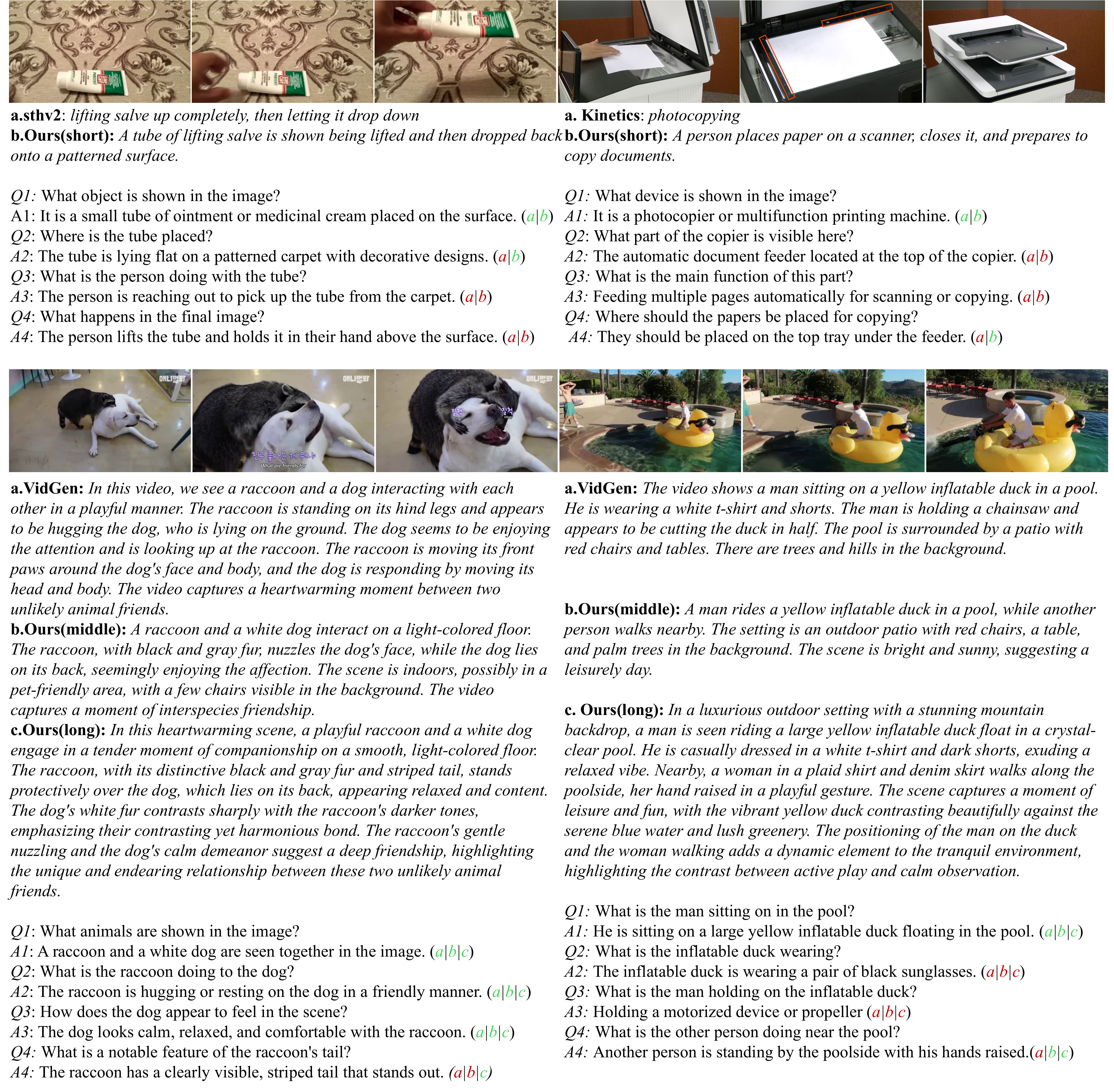}
    \captionsetup{width=0.95\linewidth}
    \caption{\textbf{Additional VQA examples comparing baseline captions with our captions across multiple datasets.} For each example, we show: baseline caption from the original dataset, and our generated caption, followed by questions and answers. Color markers (a|b) or (a|b|c) indicate which captions can correctly answer each question, demonstrating our captions' superior coverage of visual details.}
    \label{fig:supp_vqa}
\end{figure*}

To further demonstrate the richness of our captions for video question answering tasks, we provide additional qualitative examples across multiple datasets in Figure~\ref{fig:supp_vqa}. These examples showcase how our multi-granularity captions enable accurate answers to diverse question types that require fine-grained visual understanding.

For action recognition datasets (Kinetics-700, Something-Something V2), our captions provide sufficient detail to answer questions about specific actions, object interactions, and temporal sequences that are typically missing from categorical labels. For instance, in the Something-Something V2 example, our caption describes not only the lifting action but also the object's appearance and surface placement, enabling accurate responses about object type and location. For large-scale video datasets (VidGen-1M, VideoUFO), our captions capture rich contextual information including object counts, spatial arrangements, colors, and scene composition. The color-coded answer markers (a|b) or (a|b|c) clearly illustrate that original captions often fail to provide sufficient information for answering detailed questions, while our captions achieve near-complete coverage across all question types, validating their semantic richness.

\end{document}